\documentclass[11pt]{article}
\usepackage[top=1in, bottom=1in, left=1in, right=1in]{geometry}
\usepackage{geometry}
\usepackage{amsthm, soul, amsfonts}
\usepackage{amsmath, amssymb}
\usepackage{authblk, array, setspace} 
\usepackage{natbib}
\usepackage[pdftex,colorlinks=true,linkcolor=blue,citecolor=blue,urlcolor=blue,bookmarks=false,pdfpagemode=None]{hyperref}
\usepackage{enumerate}
\usepackage{graphicx}
\usepackage{float}
\usepackage{pgffor}
\newtheorem{theorem}{Theorem}[section]
\newtheorem{lemma}[theorem]{Lemma}

\newtheorem{corollary}[theorem]{Corollary}

\newtheorem{assumption}{Assumption}
\usepackage{hyperref}
\usepackage{algorithm}
\usepackage{algorithmic}

\author{Nicolo Colombo\footnote{
\tt nicolo.colombo@rhul.ac.uk}   
\ and Yang Gao\footnote{\tt yang.gao@rhul.ac.uk} \\
  {\small Department of Computer Science\\
  Royal Holloway University of London,
  Egham Hill, Egham TW20 0EX, UK 
 }}

\title{
Differentiable Architecture Pruning for Transfer Learning
}
\begin{document}
\maketitle
\begin{abstract}
We propose a new gradient-based approach 
for extracting sub-architectures
from a given large model.
Contrarily to existing pruning methods, which 
are unable to 
disentangle the network architecture and the corresponding weights, 
our architecture-pruning scheme produces 
transferable new structures  
that can be successfully retrained to solve different 
tasks.
We focus on a transfer-learning setup  
where architectures can be trained on a large data set but very 
few data points are available for fine-tuning them 
on new tasks.
We define a new gradient-based algorithm that 
trains architectures of arbitrarily low complexity
independently from the attached weights. 
Given a search space defined by an existing 
large neural model, we reformulate the architecture search 
task  
as a complexity-penalized subset-selection problem and solve
it through a two-temperature relaxation scheme.
We provide theoretical convergence guarantees and 
validate the proposed transfer-learning strategy 
on real data.


\end{abstract}

\section{Introduction}
\label{section introduction}

\paragraph{Motivation.}
Transfer learning methods aim to produce machine 
learning models that are trained on a given problem 
but perform well also on different \emph{new} tasks.
The interest in transfer learning comes from 
situations where large data sets can be 
used for solving a given \emph{training} task 
but the data associated with new tasks are too small 
to train expressive models from scratch.
The general transfer-learning strategy is to use 
the small available new data  
for \emph{adapting} a large model that has been previously 
optimized on the training data.
One option consists of keeping 
the structure of the pre-trained 
large model intact and fine-tuning 
its  
weights to solve the new task.
When very few data points are available and 
the pre-trained network is large, however, 
customized regularization strategies 
are needed to mitigate the risk of over-fitting.
Fine-tuning only a few parameters is a possible 
way out but can strongly limit the performance 
of the final model.
Another option is to prune the 
pre-trained model to reduce its complexity, 
increase \emph{transferability}, and prevent overfitting.
Existing strategies, however, focus on optimized 
models and are unable to \emph{disentangle} 
the network architecture from the attached weights.
As a consequence, the pruned version of the original model 
can hardly be interpreted as a transferable new 
\emph{architecture} 
and it is difficult to reuse it on new tasks.

\paragraph{In this paper.}
We propose a new 
Architecture Pruning (AP) approach for finding 
transferable and arbitrarily light 
sub-architectures of a given \emph{parent} 
model.
AP is not dissimilar to other existing pruning 
methods but based on a feasible approximation 
of the objective function normally used for 
Neural Architecture Search (NAS).
We conjecture that the proposed
architecture-focused objective makes it possible  
to \emph{separate} the role of 
the network architecture and 
the weights attached to it.
To validate our hypothesis, we define 
a new AP algorithm 
and use it to extract a series of 
low-complexity sub-architectures from    
state-of-the-art computer vision models with 
millions of parameters. 
The size of the obtained sub-architectures 
can be fixed \emph{a priori} and, 
in the transfer learning setup, 
adapted to the amount of data available from the new task.
Finally, we test the transferability of the obtained 
sub-architectures empirically 
by fine-tuning them on different small-size 
data sets.

\paragraph{Technical contribution.}
NAS is often formulated as 
a nested optimization problem, e.g.
\begin{align}
\label{nested optimization}
\min_{{\cal A}} {\cal L}\left({\cal A}, \ 
{\rm arg} \min_{\theta} 
{\cal L}({\cal A}, \theta, {\cal D}), \ {\cal D} \right)
\end{align}
where ${\cal L}({\cal A}, \theta, {\cal D})$ 
is a loss function 
that depends on 
the network structure, ${\cal A}$, 
the corresponding weights, $\theta$, and
the input-output data set, ${\cal D}$. 
This approach has a few practical problems: 
i) 
the architecture search space, i.e. 
the set of possible architectures to be considered, 
is ideally unbounded, 
ii) 
each new architecture, ${\cal A}'$, 
should be evaluated after solving the 
inner optimization problem, ${\rm arg} \min_{\theta} 
{\cal L}({\cal A}', \theta)$, which 
is computationally expensive,  
and 
iii) 
architectures are usually encoded as \emph{discrete} 
variables, i.e. 
$\min_{{\cal A}} {\cal L}$, 
is a high-dimensional Discrete Optimization (DO)
problem and its exact solution may require an 
exponentially large number of architecture 
evaluations.
To address these issues, we approximate 
\eqref{nested optimization} with the 
\emph{joint} mixed-DO problem
$\min_{{\cal A}, \theta} {\cal L}({\cal A}, \theta)$ 
i.e. we simultaneously search for  
an optimal architecture, ${\cal A}$ 
and the corresponding weights, 
$\theta$.  
The DO problem is then solved through 
a new \emph{two-temperature} gradient-based 
approach where 
a first approximation makes $\min_{{\cal A}, \theta} {\cal L}({\cal A}, \theta)$  
a fully continuous optimization problem and 
a second approximation is introduced 
to avoid the \emph{vanishing gradient}, 
which may prevent gradient-based iterative 
algorithm to converge when the first approximation 
is tight. 
Our scheme belongs to a class of recent 
differentiable NAS approaches, which 
are several orders of magnitude
faster than standard Genetic and 
Reinforcement Learning schemes 
(see for example 
the comparative table 
of \cite{ren2020comprehensive}) but the 
first to address the vanishing-gradient 
problem explicitly in this framework.\footnote{
To the best of our knowledge.}
Similar relaxation methods have been used in the 
binary networks literature 
(see for example \cite{bengio2013estimating}),
but this is the first time 
that similar ideas are transferred from the 
binary network literature to NAS.
Moreover, our method is provably accurate and, 
among various existing follow-ups of 
\cite{bengio2013estimating}, the 
only one that can 
be provided with quantitative convergence bounds.
\footnote{Mainly thanks to the 
novel two-temperature idea.}

We validate our hypothesis and theoretical 
findings through three sets of 
empirical experiments:
i) we compare the performance of the proposed 
two-temperature scheme and a more standard 
continuous relaxation method 
on solving a simple AP problem (on MNIST data), 
ii) we use CIFAR10 and CIFAR100 to 
test the transferability of VGG 
sub-architectures obtained through AP and 
other pruning methods
\footnote{A fair 
comparison with other NAS methods 
is non-trivial because it is not clear how to 
fix a common search space and we leave it for 
future work.
}
, and  
iii) we confirm the hypothesis of 
\cite{frankle2020pruning} that 
NAS can be often reduced to selecting 
the right layer-wise density and is quite 
insensitive to specific configurations 
of the connections.\footnote{
At least in the transfer learning setup.
}

\paragraph{Related Work.}
{\bf NAS} approaches 
\cite{zoph2016neural,liu2018darts,gaier2019weight,you2020greedynas}
look for optimal neural structures in a given search
space\footnote{Usually, the boundary of the search spaces are set 
by limiting the number of allowed neural operations, 
e.g. node or edge addition or removal.}, and  
employ \emph{weight pruning} procedures
that attempt to improve the performance of large 
(over-parameterized) networks by  removing 
the `less important' connections.
Early methods 
\cite{stanley2002evolving,zoph2016neural,real2019regularized} 
are based on expensive \emph{genetic algorithms} 
or \emph{reinforcement learning} approaches. 
More recent schemes
either design differentiable losses
\cite{liu2018darts,xie2019snas},
or use random weights to 
evaluate the performance of the architecture 
on a validation set 
\cite{gaier2019weight,pham2018efficient}. 
Unlike these methods, which search architectures by
adding new components, our method \emph{removes} redundant 
connections from  
an over-parameterized parent network.
{\bf Network pruning} approaches 
\cite{collins2014memory,Han2015LearningBW,han2015deep,Han2015LearningBW,Frankle2019TheLT,yu2019playing,frankle2020pruning}
start from pre-trained neural models and prune 
the unimportant connections 
to reduce the model size and achieve better performance. 
Contrarily to the transfer learning goals of AP, 
these methods mostly focus on single-task setups.
{\bf Network quantization and binarization}
reduce the computational
cost of neural models by using lower-precision 
weights \cite{jacob2018quantization,zhou2016dorefa}, 
or mapping and hashing similar
weights to the same value 
\cite{chen2015compressing,hu2018hashing}.
In an extreme case, the weights, 
and sometimes even the inputs, are binarized,
with
positive/negative weights 
mapped to $\pm1$
\cite{soudry2014expectation,courbariaux2016binarized,hubara2017quantized,shen2019searching,courbariaux2015binaryconnect}.
As a result, these 
methods keep all original connections, 
i.e. do not perform any architecture search.
Often used for binary network optimization, 
Straight-Through gradient Estimator (STE) algorithms 
are conceptually similar to the 
two-temperature scheme proposed here. 
STE looks at discrete variables as the output  
of possibly non-smooth and non-deterministic quantizers
which depends on real auxiliary quantization parameters 
and can be handled through (possibly approximate) gradient methods.
\cite{guo2018survey} compares a large number of recent  
deterministic and probabilistic quantization methods. 
Compared to standard discrete optimization techniques, 
STE methods are advantageous because they can be combined 
with 
stochastic iterative techniques to handle 
large models and large data sets. 
\cite{bengio2013estimating} is the first work where 
different objective functions are used 
in the backward and forward passes.
In \cite{bengio2013estimating}, 
the derivative of the quantizer is completely neglected, 
which leads to an unpredictable gradient bias and 
a non-vanishing optimization gap \cite{li2017training}.
Some theoretical control of STE
is given in \cite{yin2019understanding} under quite 
some assumptions on the objective function.
Certain convergence guarantees are obtained in 
\cite{ajanthan2019mirror} but through an 
annealing technique that should be fixed in advance.
\cite{yin2019blended} and \cite{xiong2019fast}
propose proximal gradient approaches 
where the gradient is 
guaranteed to define descent direction but  
both works lack of a full convergence proof.
\cite{hou2016loss} and \cite{uhlich2019mixed} 
define a new class 
of loss-aware binarization methods and 
are designed for taking into 
account certain `side' effects of the quantization step, 
but these methods considerably increase the size
of the parameter space. 
The proposed approach is the first to 
address explicitly the vanishing-gradient issues associated 
with the deterministic quantizers.

\section{Methods}
\label{sec:our_method}

\subsection{Problem formulation}
\label{section problem formulation}
To address the three 
NAS technical challenges 
mentioned in Section \ref{section introduction},
we
i) 
let the search space be 
the set of all sub-networks of a very large and 
general parent network defined by a given architecture, 
${\cal A}_{parent}$
and the associated weights, 
$\theta \in {\mathbb R}^{D}$, 
where 
$D = |{\cal A}_{parent}|$ is the number 
of weighted connections of ${\cal A}_{parent}$, 
ii)
we approximate \eqref{nested optimization}
with\footnote{
This enables us to find the optimal edge 
structure directly, 
without averaging over randomly sampled weights, 
as for weight-agnostic networks \cite{gaier2019weight},  
or training edge-specific real-value weights, 
as in other architecture
search methods \cite{liu2018darts,Frankle2019TheLT}. 
}
\begin{align}
\label{problem formulation general}
\min_{{\cal A} \subseteq {\cal A}_{parent}} \ \min_{ 
\theta \in {\mathbb R}^{D}}
{\cal L}({\cal A}, \theta, {\cal D}) 
\end{align}
where 
${\cal L}({\cal A}, \theta, {\cal D})$ is an arbitrary  
real-valued loss function   
and ${\cal D} = \{ (x, y) \in {\cal X} \otimes {\cal Y} \}$ 
a training data set,
iii)
we approximate the DO part of \eqref{problem formulation general}, 
i.e. the minimization over 
${\cal A} \subseteq {\cal A}_{parent}$, 
with a \emph{low-temperature} continuous relaxation 
of \eqref{problem formulation general} and 
solve it with iterative parameter updates based on a 
further 
\emph{high-temperature} approximation of the gradient.

We let the parent network be 
$F (\theta) = F({\cal A}_{parent}, \theta): {\cal X} \to {\cal Y}$.
Each sub-network of $F(\theta)$ 
is represented as a \emph{masked version} of $F(\theta)$, 
i.e. a network with architecture ${\cal A}_{parent}$ 
and masked weights
\begin{align}
 \label{weight expression}
\tilde \theta  = m \circ \theta,  \quad  m_i = \left\{ \begin{array}{cc}
     1 & i \in {\cal A}(m) \subseteq {\cal A}_{parent} \\
     0 & {\rm otherwise}
\end{array}\right.   
\end{align}
where $\circ$ is the element-wise product, 
$m \in \{ 0, 1 \}^{D}$,
$\theta \in {\mathbb R}^{D}$, 
${\cal A}(m)$ is a subset of the set of connections of 
${\cal A}_{parent}$, 
and $i=1, \dots, D$.
Equivalently, ${\cal A}(m)$ can be 
interpreted as the sub-architecture of 
${\cal A}_{parent}$ obtained by masking 
${\cal A}_{parent}$ with $m$. 
Given a data set, ${\cal D}$, 
we let the corresponding task be the prediction of 
the outputs, $y \in {\cal Y}$, given the corresponding 
input, $x \in {\cal X}$.
In the transfer learning setup, we assume we have 
access to a large data set associated with the training task, 
${\cal D}_{train}  = 
\{ (x, y) \in {\cal X}_{train} \otimes {\cal Y}_{train} \} $, 
and a small data set associated with a new task,
${\cal D}_{new}  = 
\{ (x, y) \in {\cal X}_{new} \otimes {\cal Y}_{new} \} $. 
We also assume that the two data sets may have different input-output spaces, 
i.e. we may have 
${\cal X}_{train} \neq {\cal X}_{new}$, 
${\cal Y}_{train} \neq {\cal Y}_{new}$, and  
$|{\cal D}_{train}| >> |{\cal D}_{new}|$.
The idea is to use ${\cal D}_{train}$ to 
extract a sub-architecture 
${\cal A} \subseteq {\cal A}_{parent}$, 
that can be retrained on ${\cal D}_{new}$ to solve 
the new task, i.e. the task associated ${\cal D}_{new}$. 
In our setup, this is equivalent to solve
\begin{align}
\label{AP problem formulation}
m_* = {\rm arg} \min_{m} \left(  \min_\theta {\cal L}(F(m, \theta), {\cal D}_{train})\right)
\end{align}
where 
$F(m, \theta) = F(\tilde \theta)$
${\cal L}(F(m, \theta), {\cal D}) = {\cal L}({\cal A}(m), 
\theta, {\cal D})$ with 
with $m \in \{ 0, 1 \}^{D}$.
To evaluate the \emph{transferability} of $m_*$,
we find 
\begin{align}
\label{problem transfer}
\theta_* = {\rm arg} 
\min_\theta {\cal L}(F(m_*, \theta), {\cal D}_{new})
\end{align}
and evaluate the performance of $F(\tilde \theta*) 
= F(m_*, \theta_*)$ on the new task.

\subsection{Two-temperature continuous relaxation}
\label{section two temperature}
The network mask, $m\in \{0, 1 \}^D$, is a discrete variable
and \eqref{AP problem formulation} cannot be solved with 
standard gradient-descent techniques.
Exact approaches would be intractable for any 
network of reasonable size.\footnote{
For the simple $64$-dimensional 
linear model defined in 
Section \ref{section mnist experiment}, 
the number of possible architectures, i.e. 
configurations of the discrete variable $m$,   
is  
$|{\cal P}(\{ 1, \dots, 64\})| = 2^{64}$. 
}
We propose to find possibly approximate solutions 
of \eqref{AP problem formulation} by solving 
a continuous approximation 
of \eqref{AP problem formulation} (first approximation)
through approximate gradient updates 
(second approximation).
Let $t_l >> t_s > 0$ be two constant associated with two 
temperatures ($1/t_l << 1/t_s$).
The low-temperature approximation of 
\eqref{AP problem formulation}
is obtained by replacing 
$m \in \{0, 1 \}^D $ with 
\begin{align}
\label{low temperature approximation}
v_{t_l} = \sigma(t_l w), \quad  
\quad w \in {\mathbb R}^D
\end{align}
everywhere in \eqref{AP problem formulation}.
When $t_l$ is large, the approximation is tight but 
minimizing it through gradient steps is challenging 
because the gradient of the relaxed objective with 
respect to the quantization parameter, 
$\nabla_w {\cal L}(F(v_{t_l}, \theta), {\cal D}_{train}))$, 
vanishes almost 
everywhere.\footnote{
The problem arises because  
$d/dt \sigma(t_l w)= t_l v_{t_l} (1 - v_{t_l}) \to 0$
for any $w \neq 0$.}
With high probability, naive gradient methods 
would get stuck into the exponentially-large 
flat regions of the energy landscape associated with 
{\cal L}. 
To mitigate this problem, we use a second 
high-temperature relaxation of \eqref{AP problem formulation} 
for computing an approximate version of 
the low-temperature gradient $\nabla_w {\cal L}\approx 0$.
More precisely we let 
\begin{align}
\label{gradient approximation}
&\Tilde \nabla_{w} {\cal L}= \nabla_{v_{t_l}}{\cal L}(v_{t_l},\theta, {\cal D}) \circ 
\left( t_s v_{t_s}(1 - v_{t_s})\right),  \\
& v_{t_s} = \sigma(t_s w)
\nonumber
\end{align}
The proposed scheme allows us to 
i) 
use good (low-temperature) 
approximations 
of \eqref{AP problem formulation} without 
compromising the speed and accuracy of the 
gradient-optimization process and 
ii) 
derive, under certain 
conditions, quantitative convergence bounds.

\subsection{Convergence analysis}
\label{subsec:proof}
Let $v \in [0,1]^D$ be a general version of 
\eqref{low temperature approximation}, defined as 
$v = \sigma(t w)$ for some $t > 0$ and $w \in {\mathbb R}^D$,  
$\theta \in {\mathbb R}^D$, and 
\begin{align}
{\cal L}_t(w) = |{\cal D}|^{-1} \sum_{z = (x,y) \in {\cal D}} 
\ell(z, v, \theta)), 
\end{align}
where $\ell(z; v, \theta) = \ell(z; F(v, \theta))$
is a single-sample loss, e.g. 
$\ell(z, F(v, \theta)) = (F(x; v, \theta) - y)^2$.
We assume that $\ell$ is convex 
in $v$ for all $z$, and that $\nabla_v \ell$ 
is bounded, i.e., that 
$\ell$ satisfies the assumption below.
\begin{assumption}
	\label{assumptions ell}
	For all $v, v' \in [0, 1]^D$, any possible 
	input $z$, and any parameters $\theta \in \mathbb{R}^D$, 
	$\ell(z; v, \theta)$ is differentiable with respect to $v$ and 
	\begin{align}
	\label{ass:assumption ell}
		&\max_{z,v, \theta} 
		\| \nabla_v \ell(z;v,\theta) \|^2 \leq G^2, \\
		&\ell(z;v,\theta) - \ell(z;v',\theta) \geq \nabla_{v'}
		\ell(z;v',\theta)^\intercal
		(v - v'), 
    \end{align}
	 where $G$ is a positive constant.
\end{assumption}

\noindent Under this assumption, 
we prove that gradient steps based on 
\eqref{gradient approximation} 
produce a (possibly approximate) locally optimal 
binary mask.

\begin{theorem}
\label{theorem:main}
Let $\ell$ satisfy Assumption \ref{assumptions ell} and
$\{ w_i \in {\mathbb R}^D\}_{i=1}^T$ 
be a sequence of stochastic weight 
updates defined using \eqref{gradient approximation}.
For $i=1, \dots, T$, let 
$z_i$ be a sample drawn from ${\cal D}_{train}$ uniformly 
at random and
$\alpha_i = \frac{c}{\sqrt{i}}$, where $c$ is a 
positive constant. Then 
$$
\mathbb{E}[{\cal L}_{t_l}(w_T) - 
{\cal L}_{t_l}(w^*)] 
\leq \frac{1}{c \sqrt{T}} + \\
\frac{c G^2 (1 + C) (1 + \log T)}{T},
$$
where the expectation is over the 
distribution generating the data,
$w^* = \arg \min_{w \in \mathbb{R}^d} 
{\cal L}_{t_l}(w)$, 
$G$ is defined in Eq. \eqref{ass:assumption ell} 
(with $v = \sigma(t_l w)$), 
and 
	\begin{align}
		&C = t_l t_s \left(\frac{1}{t_l t_s}  
	 - 2 g_{max}(t_l) g_{max}(t_s)
	  + \frac{t_l t_s}{16^2} \right)
	\\
		&g_{max}(t)  
    = \sigma(t M)
    (1 - \sigma(t M)), \nonumber 
	\end{align}
	with $M = \max \{ |w_i|, i=1, \dots, T \}$.
\end{theorem}

A proof is provided
in the Supplementary Material.\footnote{
For simplicity, we consider 
the convergence of updates based on 
\eqref{gradient approximation} for 
fixed $\theta$, but a full convergence proof 
can be obtained by combining 
Theorem \ref{theorem:main} with standard results 
for unconstrained SGD.}
Theorem \ref{theorem:main} and  
$\lim_{t_l \to \infty}{\mathcal L}_{t_l} = {\mathcal L}$, 
implies that a locally-optimal binary mask 
$m_* \in \{0,1\}^D$
can be obtained with high probability by letting   
$$m_* = \lim_{t_l \to \infty} \sigma(t_lw_T) 
= \mathbb m{1}[w_T > 0].$$

\section{Experiments}
\label{section experiments}

\begin{figure}[ht]
    \centering
    \includegraphics[width=.6\textwidth]{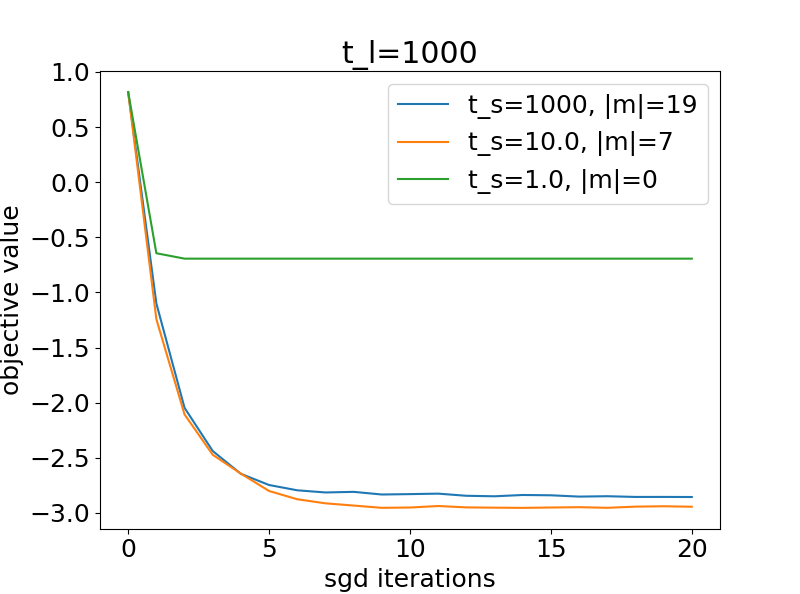}
    \caption{Convergence of the 
    gradient descent algorithm for different 
    choices of the relaxation constants, $t_s \in \{ t_l,  t_l/100, t_l/1000\}$. In the legend, 
    $|m|$ denotes the number of free parameters 
    of the final model,  ($\|m\|^2$ in 
    \eqref{mnist objective}).
    }
    \label{figure convergence}
\end{figure}

\subsection{Algorithm convergence (MNIST data, 
Figure \ref{figure convergence})}
\label{section mnist experiment}
To check the efficiency of the proposed optimization 
algorithm, we apply the 
two-temperature scheme described in \ref{sec:our_method}  
to the problem of find selecting a sparse sub-model of 
the logistic regression model
\[
F(x; \tilde \theta) = \sigma(\tilde \theta^T x) 
\]
by letting $\tilde \theta = m \circ \theta $ and 
solving 
\begin{align}
\label{mnist objective}
    &m_* = {\rm arg} \min_{m} \left( 
    \min_\theta 
    \ell(F, {\cal D}) + 
    \gamma \|m\|^2 \right) \\
    &{\cal L}(F, {\cal D})) =   
    |{\cal D}|^{-1} \sum_{(x, y) \in {\cal D}} 
    \log \left(F^y (1-F)^{1-y}\right)  
    + \gamma \|\theta\|^2  
    \nonumber 
\end{align}
We use the MNIST data set and consider the 
binary classification task of discriminating 
between images of hand-written 0s and 1s.
To evaluate the role of the relaxation 
temperatures, we replace $m$ in \eqref{mnist objective}
with $v_{t_l}$ 
defined in \eqref{low temperature approximation}, 
with $t_l=1000$, and solve the obtained low-temperature 
approximation through SGD updates based on 
\eqref{gradient approximation} where 
$t_s \in \{ t_l,  t_l/100, t_l/1000\}$.
The first case, $t_s = t_l$, is equivalent to 
a naive gradient descent approach where the 
parameter updates are computed without any gradient 
approximation.

\begin{figure}[ht]
    \centering
    \includegraphics[width=0.6\textwidth]{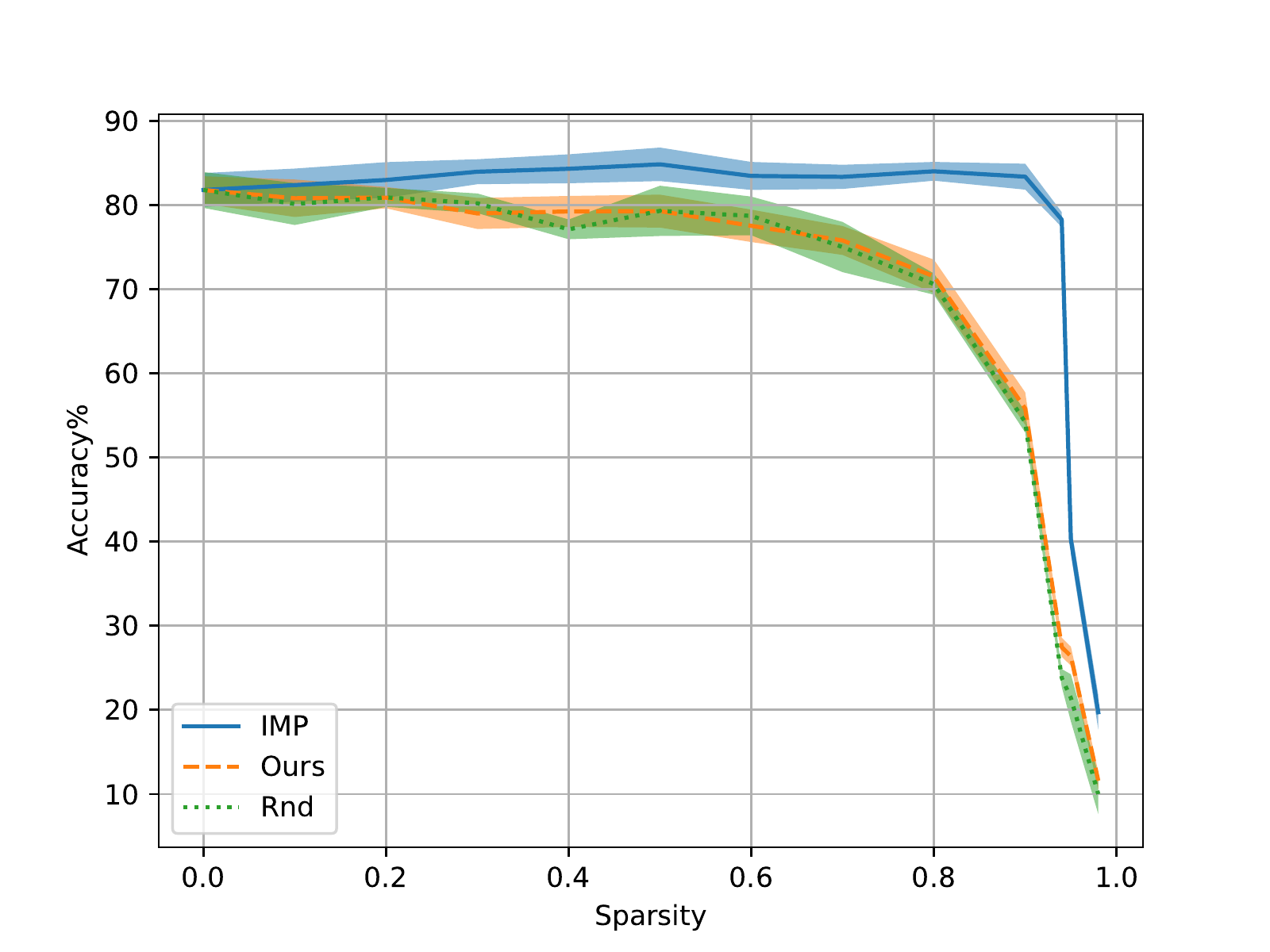}
    \caption{In-domain performance: all models
    are fine-tuned on $5k$ test images from CIFAR10 
    (same 10-class classification task used for AP).
   }
    \label{fig:in_domain}
\end{figure}

\begin{figure}[ht]
    \centering
    \includegraphics[width=0.6\textwidth]{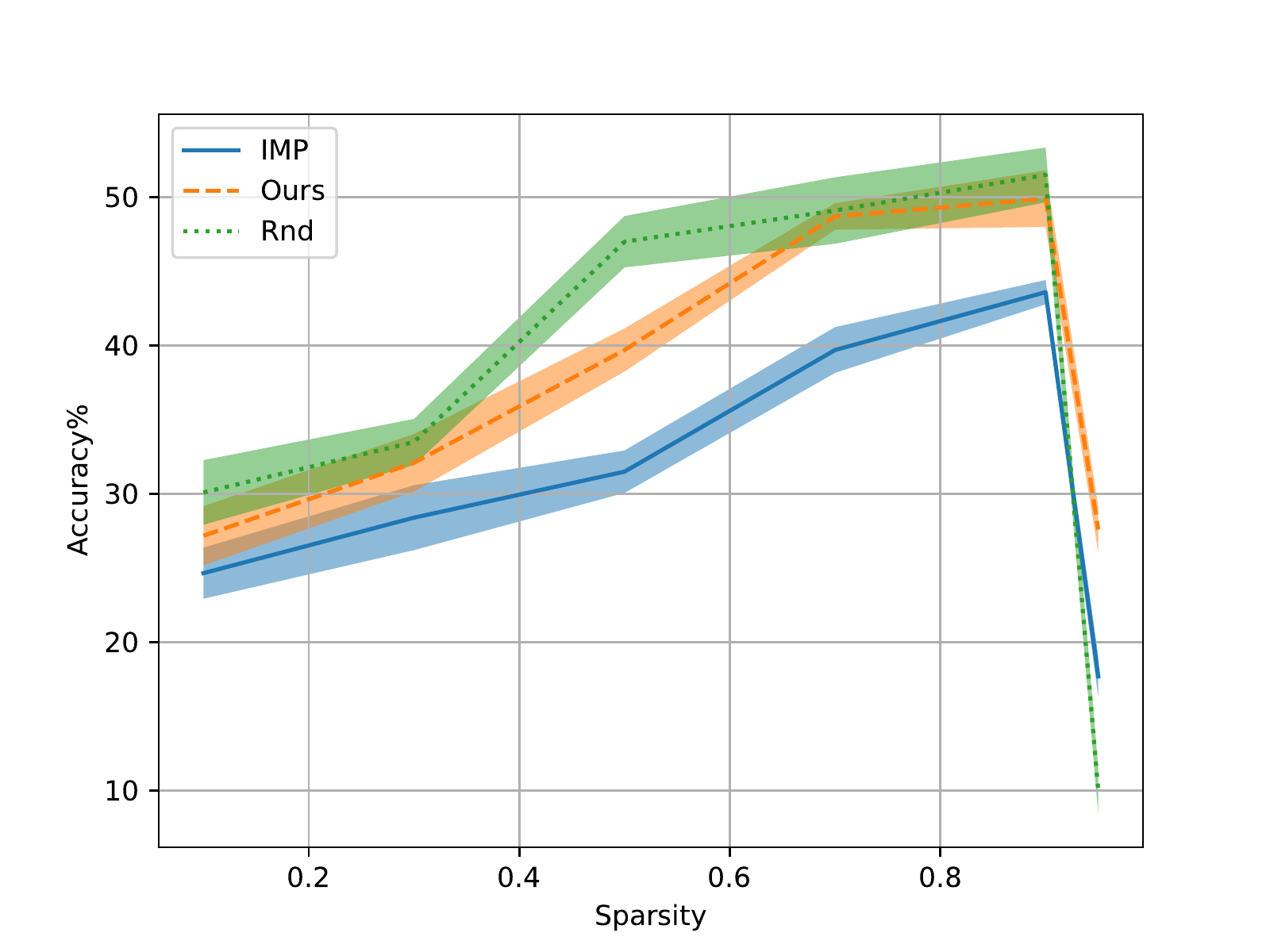}
    \caption{
    Transfer-learning performance for $|{\cal D}_{new}| = 5k$.
    }
    \label{fig:transfer_5k}
\end{figure}

\begin{figure}[ht]
    \centering
    \includegraphics[width=0.6\textwidth]{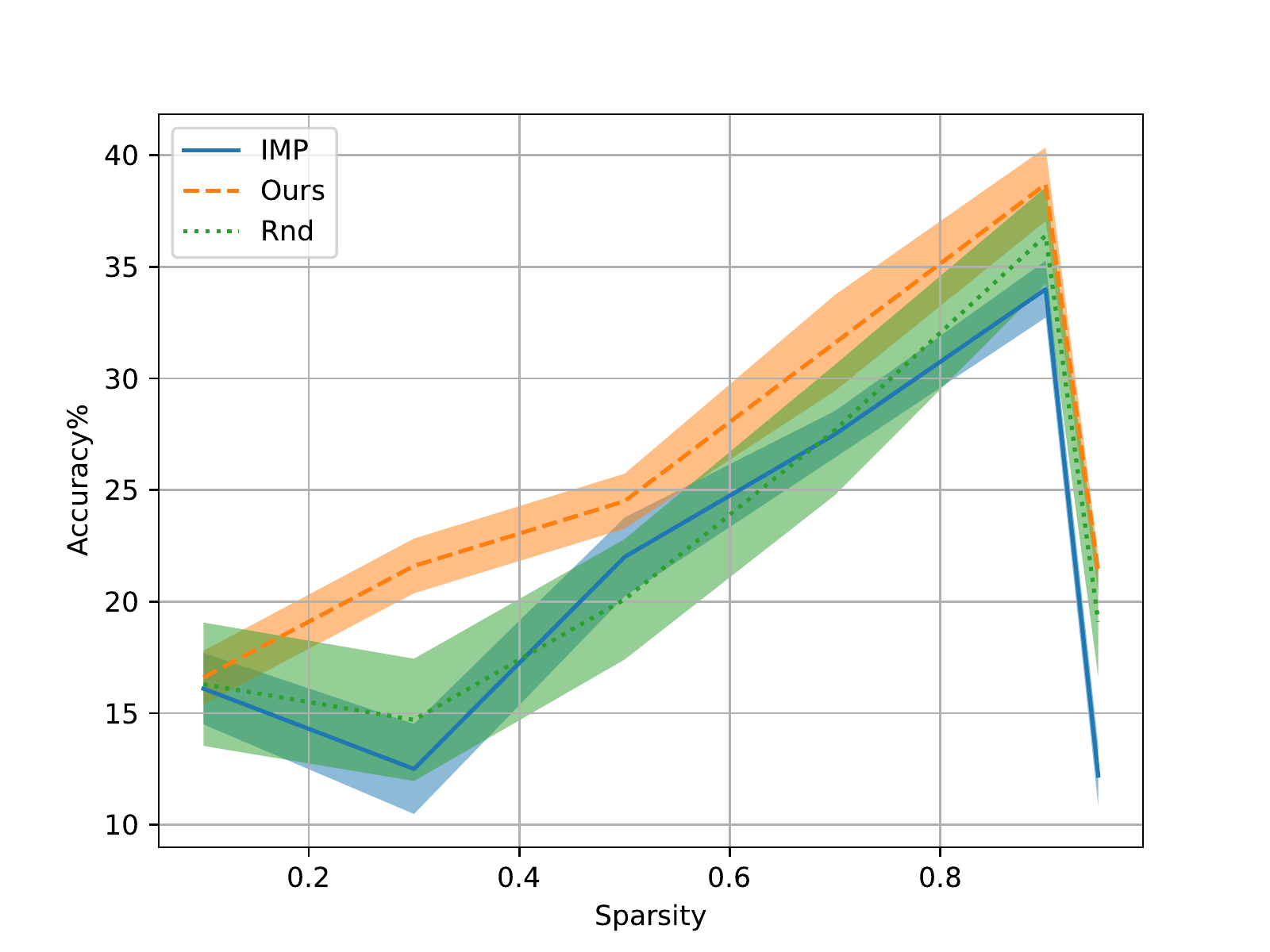}
    \caption{
    Transfer-learning performance for $|{\cal D}_{new}| = 1k$.
    $|{\cal D}_{new}| = 1k$.  
    }
    \label{fig:transfer_1k}
\end{figure}

\begin{figure}[ht]
    \centering
    \includegraphics[width=0.6\textwidth]{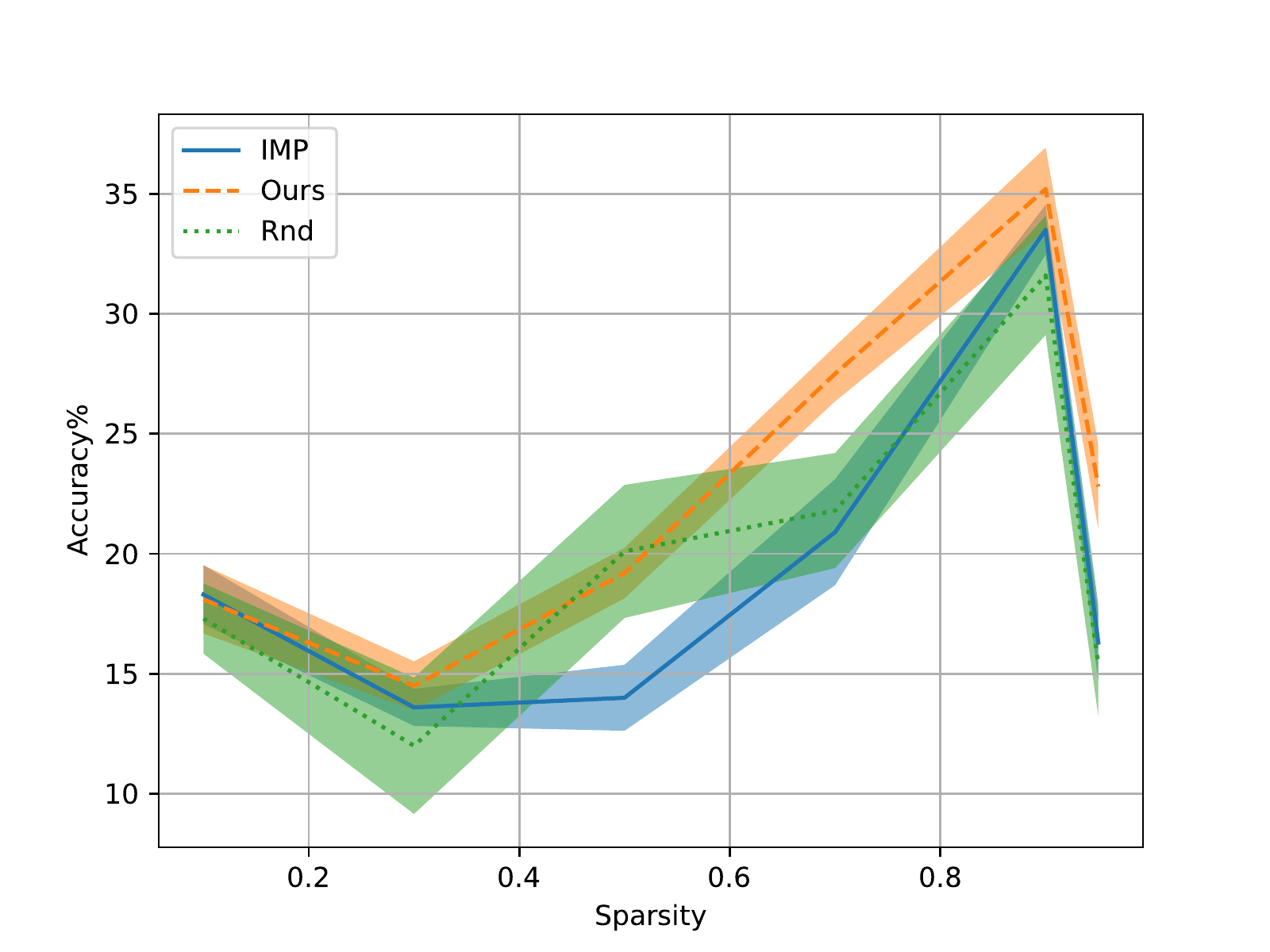}
    \caption{
    Transfer-learning performance for $|{\cal D}_{new}| = 500$.
    }
    \label{fig:transfer_500}
\end{figure}

\subsection{Transferability 
(CIFAR10 and CIFAR100 data, Figures \ref{fig:in_domain}, \ref{fig:transfer_5k}, 
\ref{fig:transfer_1k},  and 
\ref{fig:transfer_500})}
\label{section experiment transfer}
To check the scalability transfer-learning performance of AP,
we use the VGG19 \cite{simonyan2014very} model 
($D\sim 144 \ m$ free parameters) 
as parent network (see Section \ref{section problem formulation}),  
images and labels from CIFAR10 
(10 classes, 5000+1000 images per class, 
referred to as ${\cal D}_{train}$ in Section 
\ref{section problem formulation}) 
for the AP step,
and images and labels from CIFAR100 (100 classes, 
500+100 images per class, referred to as 
${\cal D}_{new}$ in Section 
\ref{section problem formulation}) 
to evaluate the obtained sub-architectures.
The VGG sub-architectures are evaluated by 
fine-tuning them on new-task 
data sets of $5k$, $1k$ $500$, $100$, $50$ images.
The new-task data sets, 
referred to as ${\cal D}_{new}$,
are obtained by randomly sub-sampling a balanced
number of images from 10 classes of CIFAR100. 
In particular, we compare the transfer-learning 
performance of VGG sub-architectures of different complexity, 
$|m| = D (1 - {sparsity})$,   
${sparsity} \in \{0.1, 0.3, 0.5, 0.9, 0.95, 0.99\}$, 
obtained with three different methods:
i) random pruning ({\tt Rnd} in the plots), 
ii) the proposed AP approach 
({\tt Ours}), and 
iii) the Iterative 
Magnitude Pruning ({\tt IMP}) scheme described in 
\cite{frankle2018lottery} (our implementation). 
In all cases, we report the average and standard 
deviations (over 5 runs) of the accuracy 
versus the models' sparsity, ${sparsity}$ defined above.

\begin{figure}[ht]
    \centering
    \includegraphics[width=0.6\textwidth]{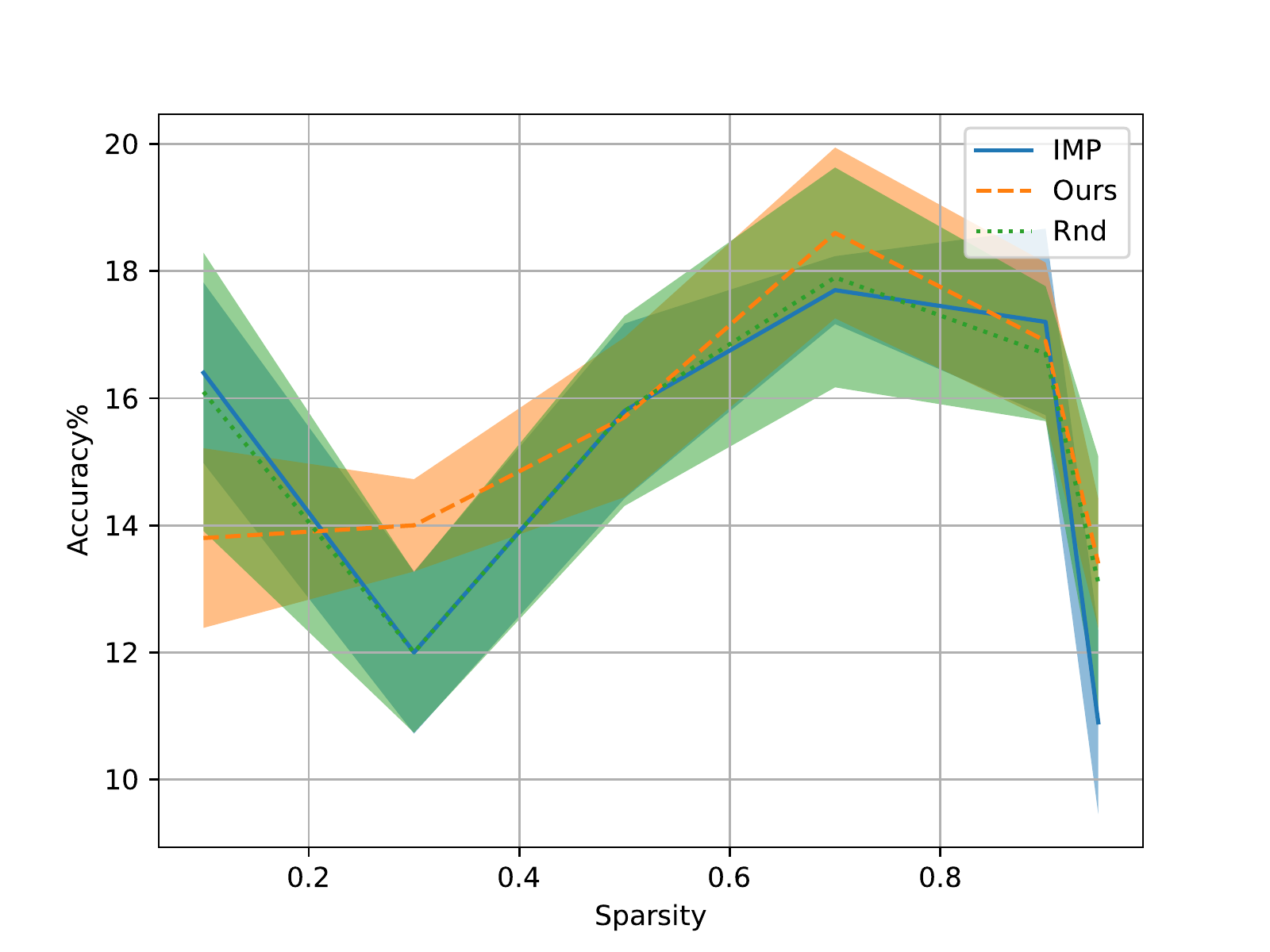}
    \caption{
    Transfer-learning performance for $|{\cal D}_{new}| = 100$.
    }
    \label{fig:transfer_100}
\end{figure}
 
\begin{figure}[ht]
    \centering
    \includegraphics[width=0.6\textwidth]{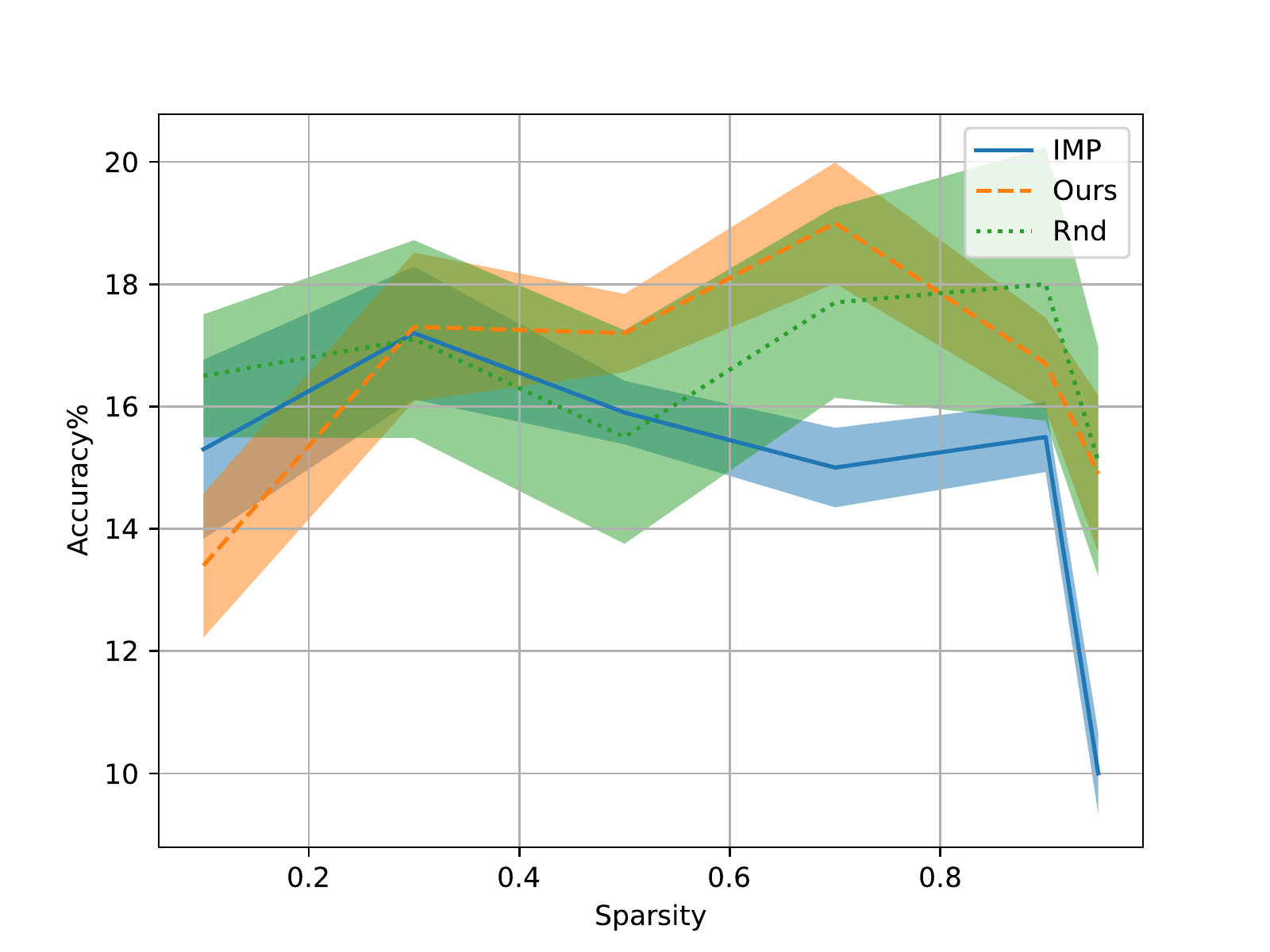}
    \caption{
    Transfer-learning performance for $|{\cal D}_{new}| = 50$.
    }
    \label{fig:transfer_50}
\end{figure}

\begin{table}[ht]
    \centering
    \begin{tabular}{|c|c c|}
    \hline 
    Sparsity & Reshuffle Ours & Reshuffle IMP \\
    \hline 
    0.1 & .378$\pm$.021 & .300$\pm$.024 \\
    0.3 & .343$\pm$.023 & .332$\pm$.022 \\
    0.5 & .429$\pm$.023 & .342$\pm$.026 \\
    0.7 & .448$\pm$.020 & .376$\pm$.025 \\
    0.9 & .497$\pm$.021 & .100$\pm$.000 \\
    0.95 & .301$\pm$.001 & .100$\pm$.000 \\
    0.99 & .100$\pm$.000 & .100$\pm$.000 \\
    \hline
    \end{tabular}
    \caption{
    Transfer-learning performance 
    after random layer-wise reshuffling
    ($|{\cal D}_{new}| = 500$).
    }
    \label{tab:shuffle}
\end{table}

\subsection{Layer-wise sparsity (CIFAR10 and 
CIFAR100 data sets, Table \ref{tab:shuffle})}
\label{section experiment density}
As noticed in previous work \cite{frankle2020pruning},
the transfer learning properties of 
neural architectures depends more 
on the layer-wise connection density
than on the specific connection configuration.
To test this hypothesis, we compare
the transfer-learning performance of 
all VGG sub-architectures upon 
layer-wise reshuffling of the 
optimized binary masks.
From each sub-architecture, 
we extract an optimal 
\emph{layer-wise} connection density, 
$1 - {sparsity}_*^{(l)} = 1^T m_*^{(l)}/D^{(l)}$,  $l=1, \dots, n_{layers}$, 
where ${sparsity}_*^{(l)}$ is the layer-wise sparsity 
and $D^{(l)}$ the total number of layer $l$ in the 
original VGG model, 
and test the transfer-learning performance of 
a random architecture with such an optimal 
layer-wise density, i.e. 
a random architecture with layer-wise binary masks 
$\tilde m^{(l)}$ satisfying
$|\tilde m^{(l)}| = D^{(l)} (1 - {sparsity}_*^{(l)})$, $l=1, \dots, n_{layers}$.

\section{Discussion}
\subsection{Results}

The proposed two-temperature approach 
improves both the speed and 
the efficiency of gradient-based algorithms 
in solving continuous relaxation of 
discrete optimization problems.
Our experiment on MNIST data (see Section 
\ref{section mnist experiment} and 
Figure \ref{figure convergence}) shows that 
a careful choice of the high-temperature 
parameter, $t_s$, 
defined in Section 
\ref{section two temperature}, may 
help the stability of the gradient updates.
Setting the high temperature to 
$t_s \sim t_l / 100$ makes a standard SGD 
algorithm 
reach a lower objective value in fewer iterations
than for $t_s = t_l$ (equivalent to using 
the exact gradient in \eqref{gradient approximation}) or 
$t_s = t_l/1000$ (higher gradient-approximation 
temperature).
The optimized models have different complexity, 
as this is implicitly controlled through 
the regularization parameter $\lambda$ in 
\eqref{mnist objective}.\footnote{To obtain
models of fixed sparsity (as in our 
transfer-learning experiments) we 
choose a suitably large $\lambda$ and stop 
updating the mask weights, $w$, 
when the target sparsity is reached.}
Choosing $t_s = t_l/1000$ 
is less efficient because the gap 
between the true gradient and its approximation 
becomes too large (in our experiments, 
it causes the AP optimization to prune all 
network connections).
These results are in line with the theoretical 
convergence bound of 
Theorem \ref{theorem:main}.

AP can be efficiently 
used to extract low-complexity and 
transferable sub-architectures 
of a given large network, e.g. VGG (see 
Section \ref{section experiment transfer}).
According to our transfer learning experiment on 
CIFAR10 and CIFAR100 (see Section 
\ref{section experiment transfer}), 
AP sub-architectures 
adapt better
than random or IMP sub-architectures of 
the same size to solve
new tasks, especially when the data available 
for retraining the networks on the new task 
is small.
When $|{\cal D}_{new}|$ is big enough, 
random sub-architectures may also perform 
well, probably because their structure is not biased by 
training on a different domain 
(Figure \ref{fig:transfer_5k}). 
AP models are consistently 
better than random when fewer than 
1000 data points are available (Figures 
\ref{fig:transfer_1k} and \ref{fig:transfer_500}).
IMP models are worse than  
AP and random models in all setups, confirming that 
IMP produces sub-architectures that are 
too strongly related to the training task 
(Figures \ref{fig:transfer_5k}, 
\ref{fig:transfer_1k},  and 
\ref{fig:transfer_500}).
When $|{\cal D}_{new}|$ is too small, e.g. 
$|{\cal D}_{new}| < 100$, all models
perform badly, which may be due to numerical 
instabilities in the fine-tuning optimization phase.
Interestingly, while lower-complexity models  
adapt better to the new domain when $|{\cal D}_{new}|$ 
is not too small (Figures \ref{fig:transfer_5k}, 
\ref{fig:transfer_1k},  and 
\ref{fig:transfer_500}), this is not true 
when $|{\cal D}_{new}|$ contains fewer than $100$ images, 
i.e. 10 images per class.
The good performance of IMP models in the 
in-domain experiment (Figure \ref{fig:in_domain})
confirms their stronger link to the original task
and a higher level of entanglement between the 
learned sub-architectures and the corresponding weights. 

The results we obtained in the 
layer-wise reshuffling experiment (Table 
\ref{tab:shuffle}) 
confirm the conjecture of
\cite{frankle2020pruning} about the 
importance of ink learning the right layer-wise 
density. 
A comparison between Table \ref{tab:shuffle} 
and Figure \ref{fig:transfer_5k} suggests 
that a good layer-wise density is what matters 
the most in making a neural architecture 
more transferable. 
Probably, the good performance of reshuffled 
and random-selected architectures also indicates that 
fine-tuning may be powerful enough to compensate 
for non-optimal architecture design when 
the number of network connections is large enough.
We should note, however, that this does not 
happen if 
i) 
the size of the new-task data is small, 
suggested by the increasing performance 
gap between AP and random models when only 1k or 500 samples 
from the new task are available
(Figures \ref{fig:transfer_1k} and \ref{fig:transfer_500}), and 
ii)
the learned sub-architectures are 
too task-specific, e.g. 
for all IMP models. 

\subsection{Directions}
Many more experimental setups can and should be tried.
For example, it would be interesting to:
\begin{itemize}
    \item test the performance of the method on the same data set 
    but for different choices parent network
    \item see if the proposed method can handle more challenging 
    transfer learning task, i.e. for less similar 
    learning and testing tasks
    \item compare with other architecture search method (this 
    is not easy as a fair comparison would require starting
    from analogous search spaces
    \item try other than random initialization in the fine-tuning 
    step, as it has been proved beneficial in similar NLP 
    transfer learning experiments (see for example \cite{chen2020lottery})
    \item study the difference between the performance of 
    the \emph{bare} architecture, i.e. with binary weights 
    taking values in $\{-1, 1 \}$, which could be used as 
    a low-memory version of the transferable models for 
    implementation on small devices) 
\end{itemize}
From the theoretical perspective, follow-up work will 
consist of applications of the proposed two-temperature
method to other discrete optimization problems.

\bibliography{refs}
\bibliographystyle{plainnat}

\newpage
\appendix
\section{Algorithms}
\subsection{Architecture-pruning algorithm}

    {\bf input:} 
    training task and data set, 
    ${\cal D}_{train}$,
    parent network architecture, 
    ${\cal A}_{parent}$, 
    low and high inverse temperatures, e.g. $t_l = 100$, 
    $t_s = t_l/10$, 
    optimization hyper-parameters, $T \in {\cal N_+}$
    $\{\alpha_i, \beta_i\}_{i=1}^T$,
    
    %
    {\bf 1.} 
    let $F(\theta) = F({\cal A}_{train}, \theta_0)$, 
    $\theta_0 \sim {\cal N}(0, \epsilon_\theta)^D$, 
    $\epsilon_\theta > 0$ be a randomly initialized version of the parent network.
    
    {\bf 2.} 
    let the training optimization problem be 
    be the low-temperature approximation of  
    \eqref{AP problem formulation} defined as  
    \begin{align}
    \label{low temperature approximated problem}
    &w_* = {\rm arg} \min_{w} \left( \min_{\theta} 
    {\cal L}(F(v_{t_l}, \theta), {\cal D}_{train})
    \right) \\
    &v_{t_l} = \sigma(t_l w)
    \end{align}
    
    {\bf 3.}
    let $w_0 \sim |{\cal N}(0, \epsilon_w)|^D$, 
    $\epsilon_w > 0$ and solve 
    \eqref{low temperature approximated problem}
    through approximate gradient updates defined by 
    \begin{align}
        &w_{i+1} = w_i - \alpha_i  
        \Tilde \nabla_v {\cal L}
        \nonumber \\
        &
        \Tilde \nabla_v 
        = \left. \nabla_v {\cal L}(z; v, \theta_i)
        \right|_{v = \sigma(t_l w_i)} \circ 
        v' \circ (1 - v')\nonumber\\
        & v' = \sigma(t_s w_i)      
        \label{sgd w}\\
        &\theta_{i+1} = \theta_i - \beta_i  
        \left. 
        \nabla_\theta {\cal L} (\sigma(t_l w_i), \theta, 
        {\cal B}_i) \right|_{\theta = \theta_i}
        \label{sgd theta}
    \end{align}
    where  $\{ {\cal B}_{i} 
    \subset {\cal D}_{train}\}_{i=1}^T $ 
    are randomly selected batches of training examples
    $z = (x, y) \in {\cal d}_{train}$
    
    {\bf output:} architecture of the sub-network obtained 
    by masking $F(\theta = 1)$ with the 
    optimized binary mask 
    \[
    m_* = {\bf 1}[w_T > 0] = \lim_{t_l \to \infty} 
    \sigma(t_l w^T)
    \]
    
\subsection{Transfer learning-evaluation algorithm}
%
    
    {\bf input:}
    new task and data set, 
    ${\cal D}_{new}$, split into 
    ${\cal D}_{retrain}$ and ${\cal D}_{test}$
    unit-weight version of the 
    parent network, $F(\theta = {\bf 1})$, 
    optimized binary mask $m_* \in \{0, 1\}^D$ 
    optimization hyper-parameters, $T \in {\mathbb N}_+$, $\{\beta_i\}_{i=1}^{T}$
    
    {\bf 1.} 
    let $F(\tilde \theta_0) = F( m_* \circ \theta_0)$, 
    $\theta_0 \sim {\cal N}(0, \epsilon_\theta)^D$, 
    $\epsilon_\theta > 0$ be a randomly initialized version of the optimized sub-architecture
    %
    
    {\bf 2.} 
    let the transfer-learning testing problem 
    be 
    \[
    \label{testing optimization problem}
    \theta* = {\rm arg} \min_{\theta} {\cal L}(F(m_*, \theta), 
    {\cal D}_{retrain})
    \]
    
    {\bf 3.} 
    solve \eqref{testing optimization problem} 
    through standard gradient descent updates 
    \[
    \theta_{i+1} = \theta_i - \beta_i m_* \circ   
        \left. 
        \nabla_\theta {\cal L} (m_*, \theta, 
        {\cal B}_i) \right|_{\theta = \theta_i}
    \]
    where $\theta_0 \sim {\cal N}(0, \epsilon_\theta)^D$, 
    $i=1, \dots, T$, 
    and $\{ {\cal B}_{i} 
    \subset {\cal D}_{test}\}_{i=1}^t $ 
    are randomly selected batches of training examples
    $z = (x, y) \in {\cal D}_{test}$
    
    {\bf output:} 
    average accuracy of $F(\tilde \theta*)$, 
    $\tilde \theta* = m* \circ \theta*$, 
    on ${\cal D}_{test}$ 
    \[
    {\rm acc}_{m_*} = |{\cal D}_{test}|^{-1} \sum_{(x, y) \in {\cal D}_{test} } {\bf 1}[{\rm arg} \max F(x; \tilde \theta*) 
    = y]
    \]
\subsection{Enforcing sparsity}
The number of active connections in the learned 
architecture is determined by the number of positive 
entries of $w_T$.
To encourage sparsity, we replace \eqref{sgd w} and 
\eqref{sgd theta} with 
\begin{align}
\label{sgd sparsity}
&w_{i+1} = w_{i} - \alpha_i \left( 
\Tilde \nabla_v {\cal L} - 2 \gamma \ 
(\mathbf{1} + w_i) \right) 
\\
& \theta_{i+1} = \theta_i - \beta_i  \left( \left.
\nabla_\theta {\cal L} (\sigma(t_l w_i), \theta, 
        {\cal B}_i) \right|_{\theta = \theta_i}
        - 2 \gamma  \ \theta_i \right) 
\nonumber 
\end{align}
where $\gamma$ is a regularisation parameter. 
This is equivalent to adding 
\begin{align}
\gamma \ 1^T \left(({\bf 1} + w)\circ({\bf 1} + w) + \theta \circ \theta \right) 
\approx \gamma \left(  \| m\|^2_0  + \|\theta \|^2_2\right)
\end{align} 
in \eqref{low temperature approximated problem}.
The mask parameters in $w$ are initialised with 
small positive values and, at each iteration, 
the penalisation term pushes them towards $-1$, so that  
sparser masks can be obtained. 
To reach the target sparsity in a 
similar number of epochs
we tune the hyper-parameter
$\gamma$.
In practice, we check the complexity of the sub-network, 
at each iteration and set $\alpha_{i+1} = 0$ 
if $1^T {\bf 1}[w_i > 0] < r_{sparsity} D$.

\section{Convergence analysis}
\subsection{Definitions}
The results presented in this appendix are 
more general than the convergence bound reported 
in the main text. 
To simplify the exposition, we use a slightly different notation. 
In particular,
\begin{itemize}
    \item $t_l$ and $t_s$ are called $M_{hard} $ and $M_{small}$
    \item $t=1, \dots, T$ 
    is used as an upper index to label the SGD epochs
    (except for the learning parameters $\alpha_t$)
    \item $m_{t_l} = \sigma(t_l \theta)$ is 
    called $v$ throughout this appendix
    \item $\ell(z; m_{t_l}, p)$ and 
    ${\cal L}_{t_l}(\theta) 
    = {\cal L}_t( \sigma(t \theta))$ of the main text are 
called $f(v, z)$ and $F(v)$
\item Theorem 1 of the main text is 
Corollary \ref{a corollary w convergence}
\item ${\cal D}_{train}$
is ${\cal D}$
\item $\nabla_v f(v, z)$ is $\nabla f(v, z)$
defined by $[\nabla f(v, z)]_i = 
\frac{\partial}{\partial v'_i} f(v', z)|_{v' = v}$
\end{itemize}  

To avoid confusion, we list of all quantities 
and conventions used throughout this technical appendix:
\begin{itemize}
\item[-] 
$d \in {\mathbf N}$: number of model parameters
\item[-]
${\cal Z} = {\cal X} \otimes {\cal Y}$: object-label space
\item[-]
$P_{Z}$: object-label distribution 
\item[-]
$Z \sim P_{Z}$: object-label random variable 
\item[-]
${\cal D} = \{z_n \in {\cal Z} | 
z_n \text{ is realization of } Z \sim P_{Z}\}_{n=1}^N $: 
training data set
\item[-]
$f: [0, 1]^d \otimes {\cal Z} \to {\mathbf R}$: single-input classification error
\item[-]
$F: [0, 1]^d \to {\mathbf R}$,  
$F(v) = \sum_{z \in {\cal D}} f(v, z) \approx 
|{\cal D}| \  {E}_{Z \sim P_Z}(f(v, Z))$: 
average classification error
\item[-]
$M_{soft}$ and $M_{hard} > 0$, such that 
$0 \leq M_{soft} \leq M_{hard}$:  
soft- and hard-binarization constants
\item[-]
$\sigma(s) = \frac{1}{1 + e^{- s}} \in [0, 1]^d$ for all 
$s \in {\mathbf R}^d$
\item[-]
$\sigma'(s): = \sigma(s)\odot (1 - \sigma(s))$,
for all $s \in {\mathbf R}^d$
\item[-]
$[\nabla g(s)]_{i} = \frac{\partial g(s')}{\partial s'_i} |_{s' = s}$ 
\item[-]
$\text{diag}(s) \in {\mathbf R}^{d \times d}$ is such that 
$[\text{diag}(s)]_{ii} = v_i$ and 
$[\text{diag}(s)]_{ij} = 0$ if $i \neq j$, $i,j = 1, \dots, d$
\end{itemize}

\subsection{Assumptions and proofs}

\begin{assumption}
	\label{a assumptions}
	$f: [0, 1]^d \otimes {\cal Z} \to {\mathbf R}$, 
	is differentiable over $[0, 1]^d$ for all $z \in {\cal Z}$ and obeys   
	\begin{align}
		&\max_{v \in [0, 1]^d, z \in {\cal Z}} 
		\| \nabla f(v, z) \|^2 \leq G^2, \\
		&f(v, z) - f(v', z) \geq \nabla f(v', z)^T (v - v'), 
\end{align}
	for all $v, v' \in [0, 1]^d$ and $z \in {\cal Z}$.
\end{assumption}

\begin{lemma}
	\label{a lemma big F}
	Let $f:[0, 1]^d \otimes {\cal Z}  \to 
{\mathbf R}$ be the function defined in Assumption \ref{a assumptions} and 
$F: [0, 1]^d \to {\mathbf R}$ be defined by 
$$F(v) = \sum_{z\in {\cal D}}f(v, z) \approx  |{\cal D}| \  E_{Z}(f(v, Z))   $$
where 
${\cal D} = \{z_n \in {\cal Z} | z_n \text{ is realization of } Z \sim P_{Z}\}_{n=1}^N $.
	Then 
	\begin{align} 
		F(v) - F(v') \geq \nabla F(v')^T (v - v'), 
	\end{align}
	for all $v, v' \in [0, 1]^d$ and $z \in {\cal X} \times {\cal Y}$.
\end{lemma}

\paragraph{Proof of Lemma \ref{a lemma big F}}
The convexity of $f$ implies the convexity of $F$ as
\begin{align}
	F(v) - F(v')  &= |{\cal D}|^{-1}\sum_{z \in {\cal D}} 
	f(v, z) - f(v', z) \\
	&\geq  
	|{\cal D}|^{-1}\sum_{z \in {\cal D}}\nabla f(v', z)^T (v - v') \\
	&= \nabla F(v')^T (v - v').
\end{align}
$\square$

\begin{lemma}
	\label{a lemma error}
	    Let $\alpha_t = \frac{c}{\sqrt{t}}$, $c > 0$, and 
	    $\theta^t \in {\mathbf R}^d$ be defined by 
	    \begin{align}
		    \label{a proof updates w}
		    \theta^{t+1} = \theta^t 
		    - \alpha_t M_{soft} \nabla f(\sigma(M_{hard} \theta^t), z_t) \odot \sigma'(M_{soft} \theta^t), 
	    \end{align}
	    where, for all 
	    $t = 1, \dots, T$, 
	    $\sigma'(s) = \sigma(s)\odot (1 + \sigma(s))$ ($s \in {\mathbf R}^d$), and $z_t$ is chosen randomly in ${\cal D}$.
	    Then, for all $t = 1, \dots, T$, $v^t = \sigma(M_{hard}\theta^t)$  obeys
	   \begin{align}
		\label{a proof updates v}
		v^{t+1}  = v^t - \alpha_t (\nabla f(v^t, z_t) + r^t), 
		\end{align}
    	where 
    	\begin{align}
    	\label{a proof r}
		    r^{t} &= \nabla f(v^t, z)  
			- M_{hard}M_{soft}  \sigma'(M_{hard} \xi^t) 
			\odot \nabla f(v^t,z) \odot \sigma'(M_{soft} \theta^t) \\
		    \xi^t &\in [\theta^t, \theta^t - 
		\alpha_t M_{soft} \nabla f(v^t, z) \odot \sigma'(M_{soft} \theta^t) ]
	\end{align}
	for any $z_t \in {\cal Z}$.
	Furthermore, for all $t = 1, \dots, T$, $r^t$, obeys 
	\begin{align}
		\| r^t \|^2 \leq G^2 C
	\end{align}
	where
	$G$ is defined in (\ref{a assumptions}) and 
	\begin{align}
		&C = M_{hard} M_{soft}	\left(\frac{1}{M_{hard} M_{soft}}  
	 - 2 g_{max}(M_{hard}) g_{max}(M_{soft})
	  + \frac{M_{hard} M_{soft}}{16^2} \right)
	\\
		&g_{max}(M)  =  \sigma(M \theta_{max})(1 - \sigma(M \theta_{max}))\\
	&\theta_{max} = \max_t |\theta^t|
	\end{align}
\end{lemma}

\paragraph{Proof of Lemma \ref{a lemma error}}
Let $\sigma_M(s) = \sigma(Ms)$, $\sigma'_M(s) = M\sigma(Ms) (1 - \sigma(Ms))$, and $v = \sigma_M(\theta)$, for any $\theta \in {\mathbf R}^d$ and $M > 0$.
Then \eqref{a proof updates w} is equivalent 
to \eqref{a proof updates v} as 
\begin{align}
	v^{t + 1} &= \sigma_{M_{hard}}\left(\theta^{t} - 
	\alpha_t \nabla f(\sigma_{M_{hard}}(\theta^t), z) 
	 \odot \sigma_{M_{soft}}'(\theta^t)\right) \\
	 &=  v^{t} 
	 - \alpha_t \sigma_{M_{hard}}'(\xi^t) \odot 
	 \nabla f(\sigma_{M_{hard}}(\theta^t), z) 
	 \odot \sigma_{M_{soft}}'(\theta^t) \\
	 & = v^{t} - \alpha_t \nabla f(v^t, z) + 
	 \alpha_t r^t \\
	 r^t &= \nabla f(v^t, z) - 
	 \text{diag}\left(\sigma_{M_{hard}}'(\xi^t) 
	 \odot \sigma_{M_{soft}}'(\theta^t) \right) 
	 \cdot \nabla f(\sigma_{M_{hard}}(\theta^t), z) \\
	 & = \nabla f(v^t, z) - 
	 \text{diag}\left(\sigma_{M_{hard}}'(\xi^t) 
	 \odot \sigma_{M_{soft}}'(\theta^t) \right) 
	 \cdot \nabla f(v^t, z)\\
	 \xi^t &\in [ \theta^t, \theta^t - \alpha_t \nabla f(v^t,z) 
	 \odot \sigma_{M_{soft}}'(\theta^t)]
\end{align}
where the first equality follows from 
$\sigma_M(a + b) - \sigma_M(a) = \sigma'_M(\xi) \ b$ for some 
$\xi \in [a, a + b]$ (mean value theorem).
For any $\theta \in {\mathbf R}$ and $z \in {\cal Z}$, 
one has 
\begin{align}
	\label{a proof bound error}
	\| r^t \|^2  &= \| \nabla f(v^t, z)\|^2 
	+ \| \text{diag}\left(\sigma_{M_{hard}}'(\xi^t) 
	\odot \sigma_{M_{soft}}'(\theta^t) \right) 
	 \cdot \nabla f(v^t, z) \|^2 \\ 
	  & \quad - 2 \nabla f(v^t,z)^T  
	  \text{diag}\left(\sigma_{M_{hard}}'(\xi^t) 
	  \odot \sigma_{M_{soft}}'(\theta^t) \right) 
	 \cdot \nabla f(v^t, z) \\
	 &\leq G^2
	 \left(1  -2 \min \sigma_{M_{hard}}'(\xi^t) 
	  \odot \sigma_{M_{soft}}'(\theta^t) 
	  + \left( \max \sigma_{M_{hard}}'(\xi^t) 
	  \odot \sigma_{M_{soft}}'(\theta^t)\right)^2 \right)\\
& \leq G^2
	 \left(1  - 2\sigma_{M_{hard}}'(\theta_{max}) \sigma_{M_{soft}}'(\theta_{max}) 
	  + \left( \frac{M_{hard} M_{soft}}{16}\right)^2 \right)\\
& \leq G^2 M_{hard} M_{soft}
	 \left(\frac{1}{M_{hard} M_{soft}}  
	 - 2 g_{max}(M_{hard}) g_{max}(M_{soft})
	  + \frac{M_{hard} M_{soft}}{16^2} \right)
	  \\
	  &  = G^2 C
\end{align}
where
$G$ is defined in Assumption \ref{a assumptions}, 
$\min a$ and $\max a$ are the smallest and largest entries 
of $a \in {\mathbf R}^d$, 
$\theta_{max} = \max_t |\theta^t |$, 
$$
g_{max}(M) = 
\sigma_{M}(\theta_{max})(1 - \sigma_{M}(\theta_{max}))
 =  \sigma(M \theta_{max})(1 - \sigma(M \theta_{max})), $$ 
and we have used 
$$\max_{s \in {\mathbf R}} \sigma_M'(s) = \sigma_M'(0) = \frac{M}{4}, 
\quad  
\min_{s \in [-a,a]} = \sigma_M'(s) = \sigma'(- a) = \sigma'(a)$$ for 
any $M > 0$, 
the Cauchy-Schwarz inequality $s^T s' \leq \|s \| \| s' \|$, and 
$s^T \text{diag}(s') \cdot s \leq (\max s') \|s\|^2$, and 
defined
$$C = M_{hard} M_{soft}	\left(\frac{1}{M_{hard} M_{soft}}  
	 - 2 g_{max}(M_{hard}) g_{max}(M_{soft})
	  + \frac{M_{hard} M_{soft}}{16^2} \right)
$$
$\square$
\begin{theorem}
	\label{a theorem convergence}
	Let $f:[0, 1]^d \otimes {\cal Z}  \to 
{\mathbf R}$ and  
$F: [0, 1]^d \to {\mathbf R}^d$ be defined as 
in Assumption \ref{a assumptions} and Lemma \ref{a lemma big F}.
Let $v^t$ and  $\alpha_t = \frac{c}{\sqrt{t}}$ 
($t=1, \dots, T$, $c > 0$)
be the sequence of weights defined in \eqref{a proof updates v}
and a sequence of decreasing learning rates and 
	$v^* = {\rm arg} \min_{v \in [0, 1]^d} f(v)$. 
	Then 
	\begin{align}
		{E}\left(
		F(v^T) - F(v^*)
		\right) 
		\leq 
		\frac{1}{c \sqrt{T}} 
	+ \frac{c G^2(1 + C) (1 + \log T)}{T}
	\end{align}
	where the expectation is over 
	$Z \sim P_{Z}$ and 
	$G$ and $C$ are defined in Lemma \ref{a lemma error}.
\end{theorem}

\paragraph{Proof of Theorem \ref{a theorem convergence}}
Assumption \eqref{a assumptions} and Lemma \eqref{a lemma big F} 
imply that $F:[0, 1]^d \to {\mathbf R}$ is a convex function 
over $[0, 1]^d$.
The first part of Lemma \ref{a lemma error} implies
that the sequence of approximated ${\mathbf R}^d$-valued  
gradient updates \eqref{a proof updates w}
can be rewritten as the sequence of approximated $[0, 1]^d$-valued 
gradient updates \eqref{a proof updates v}.
The second part of Lemma \ref{a lemma error} implies 
that the norm of all error terms in \eqref{a proof updates v} 
is bounded from above.
In particular, as each $r^t$ is multiplied by the learning rate, 
$\alpha_t = \frac{c}{t}$, 
it is possible to show that 
\eqref{a proof updates v} converges 
to a local optimum of $F:[0, 1]^d \to {\mathbf R}$.

To show that \eqref{a proof updates v} converges 
to a local optimum of $F:[0, 1]^d \to {\mathbf R}$
we follows a standard technique for proving the 
convergence of stochastic and make the further (standard) assumption 
\begin{align}
	E(r^t) = 0, \quad t = 1, \dots, T.
\end{align}
First, we let $v^t$, $r^t$ and $\alpha_t$, $t = 1, \dots, T$, 
be defined as in Lemma \ref{a lemma error}, and 
$z \in {\cal D}$ be the random sample at iteration $t + 1$.
Then 
\begin{align}
	\|v^{t + 1} - v^{*} \|^2 
	& = E\left(\|v^{t + 1} - v^{*} \|^2 \right)\\
	& = \|v^{t} - v^{*} \|^2
	- 2 \alpha_t E(\nabla f(v^t, z) - r^t)^T (v^t - v^*)
	\\ \nonumber &\quad 
	+ \alpha^2_t E\left(\| \nabla f(v^t, \tilde z) - r^t\|^2\right)\\
	& \leq \|v^{t} - v^{*} \|^2
	- 2 \alpha_t E(\nabla f(v^t, z)^T (v^t - v^*)
	+ \alpha^2_t \left(G^2 + E(\|r^t\|^2)\right)\\
	&  \leq  \|v^{t} - v^{*} \|^2
	+ 2 \alpha_t E(F(v^*) - F(v^t)) + \alpha^2_t G^2(1 + C) 
\end{align}
where $G^2$ and $C$ are defined in \eqref{a assumptions} 
and \eqref{a proof bound error}.
Rearranging the terms one obtains
\begin{align}
	\label{a F update}
	E(F(v^t) - F(v^*)) & \leq 
	\frac{\|v^{t} - v^{*} \|^2 - \|v^{t + 1} - v^{*} \|^2}{2 \alpha_t}
	+ \frac{\alpha_t}{2} G^2(1 + C) 
\end{align}
Since $F(v^T) - F(v^*)
=\frac{1}{T} \sum_{t=1}^T \left( F(v^t) - F(v^*)\right)$, 
the bound in \eqref{a F update} implies 
\begin{align}
	E\left( F(v^T) - F(v^*)\right) &  =  
	E\left( \frac{1}{T} \sum_{t=1}^T 
	\left( F(v^t) - F(v^*)\right)\right)\\
	&\leq \frac{1}{T} \sum_{t=1}^T E\left( F(v^t) - F(v^*)\right)\\ 
	&\leq \frac{1}{2 c T} \sum_{t=1}^T \left(
	\sqrt{t}(\|v^{t} - v^{*} \|^2 - \|v^{t + 1} - v^{*} \|^2) 
	+ \frac{c^2 G^2(1 + C)}{\sqrt{t}} \right) \\
	& = \frac{1}{2 c T}\left( 
	\sum_{t=1}^T (\sqrt{t+1}-\sqrt{t}
	)\|v^{t} - v^{*} \|^2 
	- T \|v^{T + 1} - v^{*} \|^2 
	+ c^2 G^2(1 + C) \sum_{t=1}^T \frac{1}{t}\right) \\
	& \leq \frac{1}{2 c T}\left( 
	\sqrt{T} \max_t \{\|v^{t} - v^{*} \|^2 \}_{t=1}^T 
	+ c^2 G^2(1 + C) (1 + \log T)\right)\\
	& \leq  \frac{1}{c \sqrt{T}} 
	+ \frac{c G^2(1 + C) (1 + \log T)}{T}
\end{align}
where the second line follows from the Jensen's inequality 
and we have used $\sum_{t=1}^T (\sqrt{t+1} - \sqrt{t}) \leq \sqrt{T}$ 
and $\sum_{t=1}^T \frac{1}{t} \leq 1 + \log T$.
$\square$

\begin{corollary}
\label{a corollary w convergence}
Let $f:[0, 1]^d \otimes {\cal Z}  \to {\mathbf R}$ and  
$F: [0, 1]^d \to {\mathbf R}^d$ be defined as 
in Assumption \ref{a assumptions} 
and Lemma \ref{a lemma big F}.
Let $\theta^t$ be a sequence of stochastic weight updates defined by  
$$
\theta^{t+1} = \theta^{t} - \alpha_t M_{soft} \nabla 
f(\sigma(M_{hard} \theta^t), z_t) \odot \sigma'(M_{soft} \theta^t)
$$
where, for all $t=1, \dots, T$,
$\alpha_t = \frac{c}{\sqrt{t}}$ ($c > 0$) and $z_t$ is chosen randomly in ${\cal D}$.
Then 
$$
E(F(\sigma(M_{hard} \theta^T)) - F(\sigma(M_{hard} \theta^*)) 
\leq \frac{1}{c \sqrt{T}} + \frac{c G^2 (1 + C) (1 + \log T)}{T}
$$
where
the expectation is over $Z \sim P_Z$,
$\theta^* = {\rm arg} \min_{\theta \in {\mathbf R}^d} F(\sigma(M_{hard} \theta)$, 
$G$ is defined in (\ref{a assumptions}), and 
	\begin{align}
		&C = M_{hard} M_{soft}	\left(\frac{1}{M_{hard} M_{soft}}  
	 - 2 g_{max}(M_{hard}) g_{max}(M_{soft})
	  + \frac{M_{hard} M_{soft}}{16^2} \right)
	\\
		&g_{max}(M)  
    = \sigma(M \theta_{max})(1 - \sigma(M \theta_{max})), 
    \quad \theta_{max} = \max_t |\theta^t|
	\end{align}
\end{corollary}

\paragraph{Proof of Corollary \ref{a corollary w convergence}}
The Corollary follows directly from Theorem \ref{a theorem convergence} because $\sigma$ is 
a strictly increasing function, which implies that  
the mappings ${\mathbf R}^d \to [0, 1]^d$ and 
$[0, 1]^d \to {\mathbf R}^d$ are both one-to-one.
In particular, for all $t=1, \dots, T$
one has $\theta^t = \frac{1}{M_{hard}} \sigma^{-1} (v^t)$
and 
\begin{align}
	\theta^* &:= \text{arg } \min_{\theta \in {\mathbf R}^d}
	F(\sigma(M_{hard} \theta))\\
	&= \frac{1}{M_{hard}} \sigma^{-1} \left(\text{arg} \min_{v \in [0, 1]^d} F(v)|_{v=\sigma(M_{hard}\theta)}\right) \\
	&= \frac{1}{M_{hard}} \sigma^{-1} (v_*)
\end{align}
with $v^* := {\rm arg} \min_{v \in [0, 1]} F(v)$.
It follows that 
$F(v^t) \to F(v*)$ implies  
$F(\sigma(M_{hard} \theta^t)) \to F(\sigma(M_{hard} \theta^*)))$.
$\square$

\end{document}